%% file: main_scnn.tex
\documentclass{article}

\usepackage{iclr2017_conference,times}
\iclrfinalcopy

\usepackage[utf8]{inputenc} % allow utf-8 input
\usepackage[T1]{fontenc}    % use 8-bit T1 fonts
\usepackage{hyperref}       % hyperlinks
\usepackage{url}            % simple URL typesetting
\usepackage{booktabs}       % professional-quality tables
\usepackage{amsfonts}       % blackboard math symbols
\usepackage{nicefrac}       % compact symbols for 1/2, etc.
\usepackage{microtype}      % microtypography
\usepackage{color}
 \usepackage{graphicx,epstopdf}  % Written by David Carlisle and Sebastian Rahtz
\usepackage{psfrag}    % Written by Craig Barratt, Michael C. Grant,
\usepackage{subfig} % Written by Steven Douglas Cochran %TABBOTCAP
\usepackage{url}       % Written by Donald Arseneau
\usepackage[hyphenbreaks]{breakurl}% only useful for latex+dvips+ps2pdf not for pdftex
\usepackage{dblfloatfix}% Written by Sigitas Tolusis
\usepackage{amsmath}   % From the American Mathematical Society
\usepackage{bm}
\usepackage{colortbl}
\usepackage{enumitem}
\usepackage{verbatim}
\usepackage{pslatex}
\usepackage{amssymb}
\usepackage{longtable}
\usepackage{listings}
\usepackage{tabularx}
\usepackage{booktabs}
\usepackage{caption}
\usepackage{afterpage}
\usepackage{multirow}
\usepackage{textcomp}
\usepackage{dcolumn}

\usepackage{array}

\iftrue
\newcommand{\jpark}[1]{{\color{blue}[\textbf{\sc Jongsoo}: \textit{#1}]}}
\newcommand{\sli}[1]{{\color{red}[\textbf{\sc Sheng}: \textit{#1}]}}
\else
\newcommand{\jpark}[1]{}
\newcommand{\sli}[1]{}
\fi

\title{
Faster CNNs with Direct Sparse Convolutions and Guided Pruning
\thanks{
This is an extended version of the paper published at International Conference on Learning Representation (ICLR) 2017 (https://openreview.net/forum?id=rJPcZ3txx).
Previous of versions of this paper have a different title ``Holistic SparseCNN: Forging the Trident of Accuracy, Speed, and Size''.
}
}

\author{
  Jongsoo Park$^1$\\
  \And
  Sheng Li$^1$\\
  \And
  Wei Wen$^2$\\
  \And
  Ping Tak Peter Tang$^1$\\
  \AND
  Hai Li$^3$\\
  \And
  Yiran Chen$^3$\\
  \And
  Pradeep Dubey$^1$\\
  \AND
  \\
  $^1$Intel Labs, $^2$Department of Electrical and Computing Engineering, University of Pittsburgh\\
  $^3$Department of Electrical and Computer Engineering, Duke University\\
  $^1$\{jongsoo.park, sheng.r.li, peter.tang, pradeep.dubey\}@intel.com, \\
  $^2$\{wew57\}@pitt.edu, $^3$\{yiran.chen, hai.li\}@duke.edu\\
}

\begin{document}

\maketitle

\input{abstract_peter}

\input{introduction_peter}

\input{scnn}
%\input{highsparsity}
%\input{holisticsparifying}
\input{experiments}

\input{related}

\input{conclusion}

\subsubsection*{Acknowledgement}
We would like to thank Yiwen Guo, Anbang Yao, and Yurong Chen for sharing the dynamic network surgery source code and their insights.
We would also like to thank Nitish Shirish Keskar for his recommendations on hyper-parameter settings.

%\newpage
%\footnotesize
%\bibliographystyle{IEEEtranS}
%\bibliographystyle{iclr2017_conference}
%\bibliography{main_scnn}

\end{document}

%% file: abstract_peter.tex
\begin{abstract}
Phenomenally successful in practical inference problems, convolutional neural networks (CNN) 
are widely deployed in mobile devices, data centers, and even supercomputers.
The number of parameters needed in CNNs, however, are often large
and undesirable. Consequently,
various methods have been developed to prune a CNN once it is trained. 
Nevertheless, the resulting CNNs offer limited benefits. While pruning the
fully connected layers reduces a CNN's size considerably, it does not improve inference speed noticeably as the compute heavy parts lie in
convolutions. Pruning CNNs in a way that increase inference speed often imposes
specific sparsity structures, thus limiting the achievable sparsity levels.

We present a method to realize simultaneously size economy 
and speed improvement while pruning CNNs.
Paramount to our success is an efficient general sparse-with-dense matrix
multiplication implementation that is applicable to convolution of feature maps
with kernels of arbitrary sparsity patterns.
Complementing this, we developed a performance model that predicts sweet spots
of sparsity levels for different layers and on different computer
architectures.
Together, these two allow us to demonstrate 3.1--7.3$\times$ convolution
speedups over dense convolution in AlexNet, on Intel Atom, Xeon, and Xeon Phi
processors, spanning the spectrum from mobile devices to supercomputers.
We also open source our project at https://github.com/IntelLabs/SkimCaffe.
\end{abstract}

%% file: introduction_peter.tex
\section{Introduction}
\label{sec:introduction}

Due to the success of deep neural networks in a broad set of practical and even critical artificial
intelligence tasks, they are now widely deployed in a spectrum of
platforms: smart phones, autonomous cars, data center servers, and even supercomputers.
While suitably designed and trained CNNs can be powerful, they are often large -- requiring
many parameters (e.g., the celebrated AlexNet~\citep{alexnet} has 60 millions).
That large neural network models incur cost in terms of memory, energy, and inference speed
is easy to see.

This motivated a line of research
(\cite{prune,deep-compression,dns,fact_lecun},
to name a few) that tries to prune the parameters after a CNN design is trained and proved
useful.
A common thread is to post-process a trained CNN. Post-processing may consist
of
retraining with sparsity inducing regularization or of approximating tensors of parameters via
tensor factorization. These methods reduce the size of CNNs significantly while preserving
inference accuracy. Nevertheless, the inference speed
gains in pruned networks is not nearly as impressive as the size reduction.
In this sense, the benefits of CNN pruning seem not fully realized.

While seemingly unintuitive, that the significantly pruned CNNs run
not nearly as significantly
faster can be easily explained.
First, fully connected ({\tt fc}) layers usually contain the bulk of the
parameters while convolutional ({\tt conv}) layers consume the bulk of
computation time.
This property shows that reducing the size of just the {\tt fc} layers will
readily lead to meaningful reduction in size as in \cite{deep-compression,dns};
but little speed improvement.
%Research work such as~\citep{deep-compression,dns} exemplifies
%successful size reduction by targeting the {\tt fc} layers seriously.
%Reduction of {\tt conv} layer sizes are addressed but not the subject of main focus,
%although sometimes reduced FLOP counts in the pruned {\tt conv} layers are reported.

The crux of speed improvement thus lie in actual fast convolution of sparse kernels
with
feature maps (not just floating-point operations reduction), which is a challenging problem. It is well known in the field of
numerical linear algebra that the performance of sparse matrix operations is
typically memory bandwidth bound.
Direct application of the sparse matrix operations to compute the
{\tt conv} layers when the kernels are sparse will likely result in
sub-optimal speed gains.
This concern on low efficiency of sparse operations is also discussed in the
design of GoogLeNet~\citep{googlenet}.
%\cite{scnn} has to highly specialize an SpMV like approach
%and use automated source code generation for efficient execution.
We will term methods
that work directly with sparse data structures ``sparse methods.''
Alternative to sparse methods, ``dense methods'' gather data in a way that allow
the actual convolution be performed by dense linear algebra functions such
as GEMM.
An example is found in~\citep{scnn_lebedev,sscnn}
which produces
some group-wise sparsity patterns that facilitate the use of existing and highly tuned
dense matrix computation library functions to perform the convolutions.
%There
%are inevitable performance overheads in preparing the sparse convolutions to
%be used this way.
However, imposing sparsity patterns limits the sparsity level
that would otherwise be achievable had arbitrary patterns been allowed. We note that
high compression in the {\tt conv} layers are gaining importance as these layers
consist a much larger percentage of parameters in recent
networks such as GoogLeNet~\citep{googlenet} and ResNet~\citep{resnet}.

We view sparse methods differently. Convolutions in CNNs involve multiple channels and
thus offer much higher data reuse than typical sparse matrix operations in
scientific computing. Specifically, we present
a highly efficient direct sparse convolution design
formulated as sparse-matrix-dense-matrix multiplication with the dense matrix columns
generated on-the-fly from a single column vector. In addition to being highly
efficient, this sparse convolution design is friendly with convolution
kernels with arbitrary sparsity patterns. We call this {\it element-wise sparsity}
to distinguish it from {\it group-wise sparsity} mentioned previously. As shown
later on, accepting element-wise sparsity significantly increases the achievable
sparsity level.

Complementing our sparse convolution design, we formulate a performance model to
elucidate when and how best to use sparse convolutions on different computer architectures and at different CNN layers. Our formulation follows the roofline
model~\citep{roofline}. In particular, our model suggests (correctly)
that sparse convolution can improve inference speed even with a moderate
sparsity level of around 70\%.
In addition, the model provides upper and
lower bounds of sparsity levels that can contribute to speed improvements.
Sparsity higher than the upper bound offer no further speed improvement; and
sparsity
lower than the lower bound can in fact slow down inference rather than accelerating it.

Combining the sparse convolution design with the performance model allows us
to prune a CNN in a co-design manner, with our proposed new pruning algorithm---Guided Sparsity Learning (GSL). As illustrated later, we can adjust
sparsity targets precisely at different layers so as to maximize inference speed, best preserve accuracy, and minimize
a network's size. In some cases, particular layers are identified as
best not pruned at all due to no potential speedups, leaving them
unchanged gives other layers more room for
gainful pruning in terms of size and speed.

Our paper makes the following contributions:\vspace{-0.2cm}
\begin{itemize}[leftmargin=*]
 \item A high performance sparse convolution design that takes advantage of
 arbitrary sparsity patterns and outperforms dense convolution even with a
 moderate sparsity.
 %Our design is formulated with sparse matrix times dense matrix with the
 %dense matrix columns generated on-the-fly from a single column vector.
 %This allows (1) avoiding the bandwidth-wasting lowering step, which is
 %particularly important for sparse convolution with low arithmetic intensity,
 %and (2) cache blocking to exploit reuse across multiple channels.
 \item A general performance model that (1) projects speedups over dense
 convolutions on varying level/types of sparsity and different computing platforms and (2) provides training guidelines for precisely targeting
 layers and sparsity ranges with potential to accelerate inference. %that can  useful ranges of different types  .
 %We use the roofline performance model that accounts for the peak arithmetic
 %operations per second and achievable memory bandwidth, which has been a
 %standard practice in high performance computing domain~\cite{roofline}.
 %Applying our model to the state-of-the-art results demonstrates sparse CNN
 %exploiting element-wise sparsity should outperform the methods with group-wise
 %sparsity.
 \item Guided Sparsity Learning (GSL), the first pruning algorithm fusing the awareness of speedup potential into sparsity learning;  and its application to AlexNet and GoogLeNet.
 In particular, in GoogLeNet, we prune out more than 80\% of parameters of all 5$\times$5/3$\times$3 {\tt conv} layers and {\tt fc} layers with no accuracy drop.
 %    our new pruning algorithm, Training guidelines based on the performance model for targeting the
% useful range of sparsity
 %We also balance sparsity among multiple layers to optimize accuracy, speed,
 %and model size together.
 %We apply the guidelines to AlexNet and GoogLeNet to obtain high sparsity of
 %XX\% and YY\%, respectively.
 %To the best of our knowledge, this is the first time to obtain high sparsity
 %in deep networks like GoogLeNet. \jpark{need to make sure this by more
 %literature survey}.
 %\item Comparison with the state-of-the-art method that focuses on
 %fully-connected layers~\cite{dns} and that uses group-wise
 %sparsity~\cite{sscnn}.
 %Based on the best published results and our own experiments, we show that
 %our method provides faster inference speed. \jpark{need numbers here}.
 \item An optimized sparse convolution implementation
 (http://github.com/IntelLabs/SkimCaffe) that provides 7.3$\times$,
 3.4$\times$, 3.1$\times$ speedups of convolution layers in AlexNet over dense
 methods on Intel Atom, Xeon, and Knights Landing processors, respectively,
 with no accuracy drop.
 In particular, this paper is one of the first evaluations of Xeon Phi
 processors on deep learning algorithms.
\end{itemize}

The rest of the paper is organized as follows.
Section~\ref{sec:sparseCNN} presents the details of our sparse convolution design,
formulation of the performance model, the Guided Sparsity Learning (GSL) pruning algorithm, and how they are combined to prune and accelerate CNNs. Section~\ref{sec:experiments} demonstrates the effectiveness
of these developments on AlexNet and GoogLeNet on a variety of platforms.
Section~\ref{sec:related} discusses related works and review the state of the
art. Section~\ref{sec:conclusion} concludes and outlines a few next-steps.

%% file: scnn.tex
\section{Going Faster with Direct Sparse Convolution, Performance Model, and Guided Pruning}
\label{sec:sparseCNN}
\newcommand{\tens}[1]{\ensuremath{\bm{\mathcal{#1}}}}

%While sparse methods have demonstrated potential to accelerate inference speed of CNNs,  main challenges still prevent sparse methods from achieving the full potential. The first and foremost challenge is the lack of efficient sparse convolution algorithms.

As explained previously, prunning CNN models does not yet benefit inference
speed as much as model size reduction.
This section first presents our efficient direct sparse convolution design that
remedies this situation significantly.
We then develop a performance model that projects speedup over different
sparsity levels and on different processor architectures.
The model guides our speedup-aware training method, Guided Sparstiy Learning
(GSL).
%GSEL precisely targets the layers with high speedup potential for active
%pruning to further improve inference speed.

\begin{figure*}
\begin{minipage}{.48\columnwidth}
\includegraphics[width=\columnwidth]{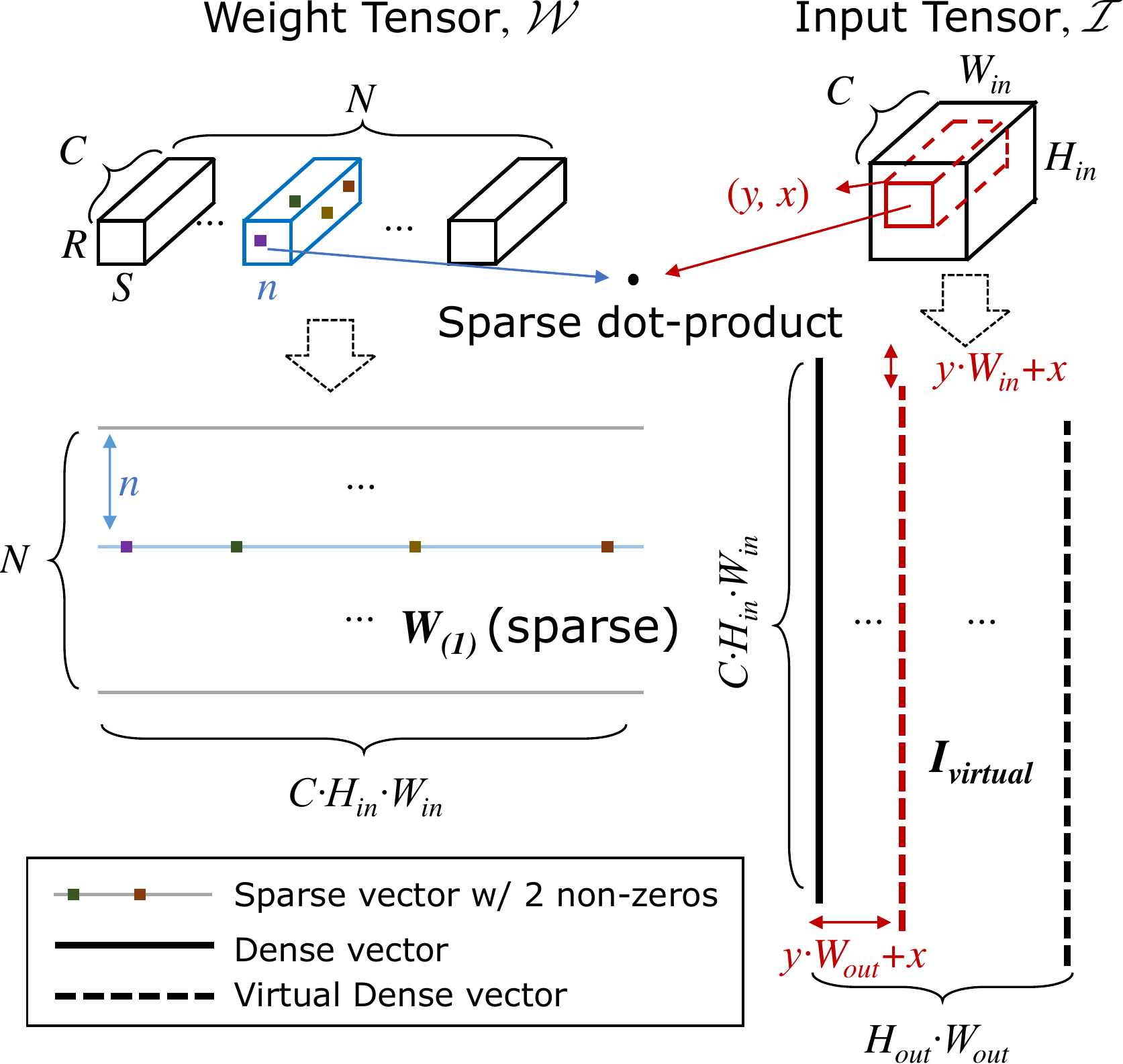}
\caption{\small Conceptual view of the direct sparse convolution algorithm.
Computation of output value at $(y,x)$th position of $n$th output channel is
highlighted.
}
\label{fig:direct_sconv}
\end{minipage}
\hfill
\begin{minipage}{.45\columnwidth}
\begin{footnotesize}
\begin{verbatim}
for each output channel n {
 for j in [W.rowptr[n], W.rowptr[n+1]) {
  off = W.colidx[j]; coeff = W.value[j]
  for (int y = 0; y < H_OUT; ++y) {
   for (int x = 0; x < W_OUT; ++x) {
    out[n][y][x] += coeff*in[off+f(0,y,x)]
   }
  }
 }
}
\end{verbatim}
\caption{\small Sparse convolution pseudo code.
Matrix $\mathbf{W}$ has {\em compressed sparse row} (CSR) format, where {\tt
rowptr[n]} points to the first non-zero weight of $n$th output channel.
For the $j$th non-zero weight at $(n,c,r,s)$, {\tt W.colidx[j]} contains the offset
to $(c,r,s)$th element of tensor {\tt in}, which is pre-computed by layout
function as $f(c,r,s)$.
If {\tt in} has CHW format, $f(c,r,s)=(c H_{in} + r)W_{in} + s$.
The ``virtual'' dense matrix is formed on-the-fly by shifting {\tt
in} by $(0,y,x)$.}
\label{fig:code}
\end{footnotesize}
\end{minipage}
\end{figure*}

\subsection{Direct Sparse Convolution}
\label{subsec:direct_sparse}
%\subsection{Why Direct Sparse Convolution?}

A sparse convolution for the all output positions across all output channels
can be eventually considered as a \emph{virtual} sparse-matrix-dense-matrix
multiplication (SpMDM), as described in the following.
Consider
%\footnote{Mini-batch size of one is used for easy illustration. Adding multiple images inside a mini-batch to this description is straightforward}
a bank of $N$ filters each with size $R\times S$ against an
$H_{in}\times W_{in}$ feature with $C$ input channels.
We denote the filter bank as a 4-mode tensor $\bm{\mathcal{W}}$ with size $N\times C
\times R\times S$, the input feature as a 3-mode tensor $\tens{I}$ with size
$C\times H_{in}\times W_{in}$, and the output feature as a 3-mode tensor
$\tens{O}$ with size $N\times H_{out}\times W_{out}$.
The output value at $(y, x)$th position of $n$th output channel is computed by
\begin{align}
\tens{O}(n,y,x) = \sum_{c=0}^{C-1} \sum_{r=0}^{R-1} \sum_{s=0}^{S-1}
\tens{W}(n,c,r,s)\tens{I}(c,y+r,x+s),
\label{eq:conv}
\end{align}
which is a dot-product of two 3D tensors as shown in
Figure~\ref{fig:direct_sconv}.
This can be treated as a vector dot-product by: first vectorizing the 3D
subtensor of $\tens{W}$ corresponding to the $n$th output channel, then vectorizing
$\tens{I}$ (denoted as $vec(\tens{I})$), and finally stretching the first vector to
match the dimension of two vectors.
When $\tens{W}$ is sparse, then this vector dot-product becomes a
sparse-vector-dense-vector dot-product.
Consider flattening dimensions except the first one of $\tens{W}$ into a sparse
matrix $\mathbf{W}_{(1)}$ (i.e.  mode-1 matricization of $\tens{W}$ as in
\cite{kolda}), with its row vectors stretched to match the dimension of
$vec(\tens{I})$.
$\tens{O}(n, y, x)$ is then the dot-product between the $n$th row of
$\mathbf{W}_{(1)}$ and $vec(\tens{I})$.
Subsequently, the values at the same given $(y, x)$th position of all $N$ output
channels can be computed collectively as a sparse-matrix-dense-vector multiplication
(SpMV):
\begin{align}
\mathbf{O}_{(1)}\left(:,yW_{out}+x\right) = \mathbf{W}_{(1)} \cdot vec\left(\tens{I}_{y,x}\right),
\end{align}
where $\tens{I}_{y,x}$ denotes the tensor $\tens{I}$ with its last two
dimensions shifted by $(y, x)$.
The values at different output positions can be computed as a
sparse-matrix-dense-matrix multiplication (SpMDM), where the columns of the dense
matrix are actually the same vector $vec(\tens{I})$ but with different offsets.
In order to save bandwidth usage, we operate with a {\em virtual} dense matrix,
$\mathbf{I_{virtual}}$, where its columns are generated on the fly by
adjusting indices through which we access $vec(\tens{I})$.

Using the virtual dense matrix essentially skips the lowering step used in
standard frameworks such Caffe, and, therefore, we call our method {\em direct
sparse convolution}, to distinguish it from sparse convolution with lowering
such as the method used in~\cite{scnn}.
The lowering approach replicates the input feature multiple times,
significantly reducing arithmetic intensity.
The lowering process has demonstrated overhead for dense convolution as in
\cite{caffe-con-troll, convbenchmark}, and is particularly problematic for sparse
convolution with intensity already lower than its dense counter part.
Figure~\ref{fig:arithmetic-intensity} demonstrates the advantage of
our direct sparse convolution, using the performance model that will be
developed Section~\ref{subsec:model}, where our direct sparse convolution
significantly outperforms lowering-based methods at a high level of sparsity.

Even though direct sparse convolution may seem conceptually more
complicated than the usual SpMDM or the lowering-based methods, it
can be concisely expressed in the pseudo code shown in
Figure~\ref{fig:code}.
To decouple from a specific layout of tensor $\tens{I}$, the pseudo code uses
layout function $f$ such that $f(c,y,x)$ maps to the offset corresponding to
$(c, y, x)$th element of $\tens{I}$ (we assume $f(c,y+r,x+s) = f(c,y,x) +
f(0,r,s)$).
For example, in CHW layout, $f(c,y,x) = (c\cdot H_{in} + y)W_{in} + x$.

In convolutional layers in CNN, an input channel is reused against multiple output
channels and vice versa, and there is also ample reuse out of an input
channel due to overlapping between dot-products, especially for a large
filter size and a small stride.
Therefore, the arithmetic intensity of sparse convolution can be
significantly higher than typical sparse-matrix computations such as SpMV,
thus leading to high compute efficiency.
Our optimized implementation\footnote{\scriptsize
https://github.com/IntelLabs/SkimCaffe/blob/intel\_scnn/include/caffe/util/sconv.hpp
}
fully takes advantage of the reuse, applying loop tiling to both input and
output channels, with column blocking~\citep{csb} applied to the weight sparse
matrix.
SIMDification and register blocking optimizations are also applied to the
{\tt y} and {\tt x} loops in the pseudo code.
Non-contiguous indirect access (i.e. gather) is another overhead of typical
sparse-matrix computations.
However, as shown in our pseudo code, the values read from {\tt colidx} and {\tt
value} arrays of a sparse matrix are reused $H_{out}\cdot W_{out}$ times.
The access to tensor {\tt in} is also contiguous as long as the tensor's elements with
contiguous $x$ or $y$ values are stored contiguously, which is a common case as in the CHW format.

Even though a bulk of computation belongs to convolution layers (hence the
focus of this paper), we also briefly discuss exploiting sparsity in fully
connected layers.
%\sli{enabling FC together, better than perforation}
Exploiting sparsity in fully connected layers is actually simpler than
convolution layers, because fully connected layers are implemented as {\tt GEMM} and
we can leverage work done before on sparse matrix and dense matrix multiplication (SpMDM).
% (this is actually similar to what would be implemented for convolution
%sparse convolution with lowering).
Similarly to sparse convolutions, the arithmetic intensity of SpMDM decreases
with higher sparsity, and its actual FLOP/s is lower than that of {\tt GEMM}.
Therefore, we also have a range of useful sparsity that can guide training for
balanced accuracy-speed-size trade-offs.
We apply optimizations similar to the ones applied to direct sparse convolution
such as loop tiling and register blocking.
%Thus, we realize speedups higher ({\bf TODO: put numbers}) than the SpMDM
%results reported in \cite{scnn} as will be shown in Section~\ref{sec:experiments}.

We briefly discuss how the sparse matrices arising from CNN differ
from the ones in scientific computing or graph analytics~\cite{uf}.
Sparse matrices in scientific computing typically have banded non-zero patterns
(i.e., a long diameter in their graph interpretations) that lead to high
temporal locality.
We observe however that matrices in sparse CNN do not exhibit such banded
non-zero patterns and reordering algorithms such as reverse Cuthill
McKee~\cite{rcm} improve the locality little.
Therefore, existing sparse linear algebra library routines like the ones in
Intel MKL that have been optimized for scientific matrices do not provide the
best performance, hence need for a custom implementation like ours with loop
tiling and register blocking.
Graph analytics is another area where sparse matrix is extensively used.
Sparse matrices in graph analytics are much bigger and often exhibit
``power-law'' distribution that desires a different data structure like doubly
compressed sparse column~\cite{hypersparse}.

We observe input features also have high sparsity (up to 85\%) in FC layers in
AlexNet that motivates using sparse-matrix sparse-matrix multiplication
(SpGEMM).
However, we evaluate that its performance is lower than SpMDM.
A challenge is that the size of output matrix is unknown a priori, requiring
two passes over the input matrices for parallelization (the first pass to
determine the number of non-zeros per output matrix row and the second pass for
the actual computation)~\cite{Gustavson}.
There are ways to work around, for example by outputting a dense matrix or by
having a separate output sparse matrix per thread, but they are not without
their own overheads.
%Another challenge is that SpGEMM involves frequent indirect writes (scatters)
%which are typically more inefficient than indirect reads (gathers).
As a result, even the state-of-the art SpGEMM implementation on a recent Xeon
processor that is significantly faster than its GPU counterparts do not achieve
more than 20 GFLOP/s~\cite{spgemm}.
In addition, the sparsity of FC layer inputs highly depends on activation
functions.
Even though ReLU with zero slope in negative side is currently a popular choice
of activation function resulting in high sparsity, we cannot be certain that
this will continue.
Still, it is interesting to see if these challenges can be overcome and SpGEMM
can further improve the performance of FC layers in sparse CNN.

\subsection{Analytical Performance Modeling: Projections on Sparse Convolution Speedup and
Guidelines on Useful Sparsity Range}

\label{subsec:model}

The performance of sparse convolution depends highly on the sparsity level of the weight
tensor.
This section develops a performance model to determine the appropriate target
sparsity range for pruning and to project theoretical speedup for any given sparsity,
using the roofline model~\citep{roofline}.

We denote the floating-point operations required in a convolution as $C$ (in FLOP),
the size of input and output activation tensors as $S_A$ (in Bytes), and the
size of weight tensor as $S_W$, all without considering sparsity.
We denote the density of non-zero in filters as
$x$ (the lower the $x$, the higher the sparsity of weight tensor), the compute
capability of processor as $F$ (in FLOP/s), and the memory bandwidth as $B$ (in
B/s).
With these parameters, the time for dense convolution ($t_\text{dense}$), the time for sparse convolution bound by
compute ($t_\text{sparse\_compute}$) and by bandwidth ($t_\text{sparse\_bw}$), and theoretical speedup can be modeled as follows (we
assume dense convolution is not bandwidth bound):
\[
t_\text{dense} = \frac{C}{F},\ \
t_\text{sparse\_compute} = \frac{\alpha xC}{F},\ \
t_\text{sparse\_bw} = \frac{S_A + \beta xS_W}{B},\ \
\text{speedup} = \frac{t_\text{dense}}{max(t_\text{sparse\_compute}, t_\text{sparse\_bw})},
\]
where $\alpha$ and $\beta$ denote the compute and storage overheads of sparse
representations, respectively.
We observe $\alpha \sim 3$ on a Xeon E5-2697 v4 processor, and $\beta$ is
typically 2 (in compressed sparse row representation, we need 4B column
index for each 4B single-precision floating point non-zero value).

\begin{figure}
\subfloat[b][Xeon E5-2697 v4]{
\includegraphics[width=.56\columnwidth]{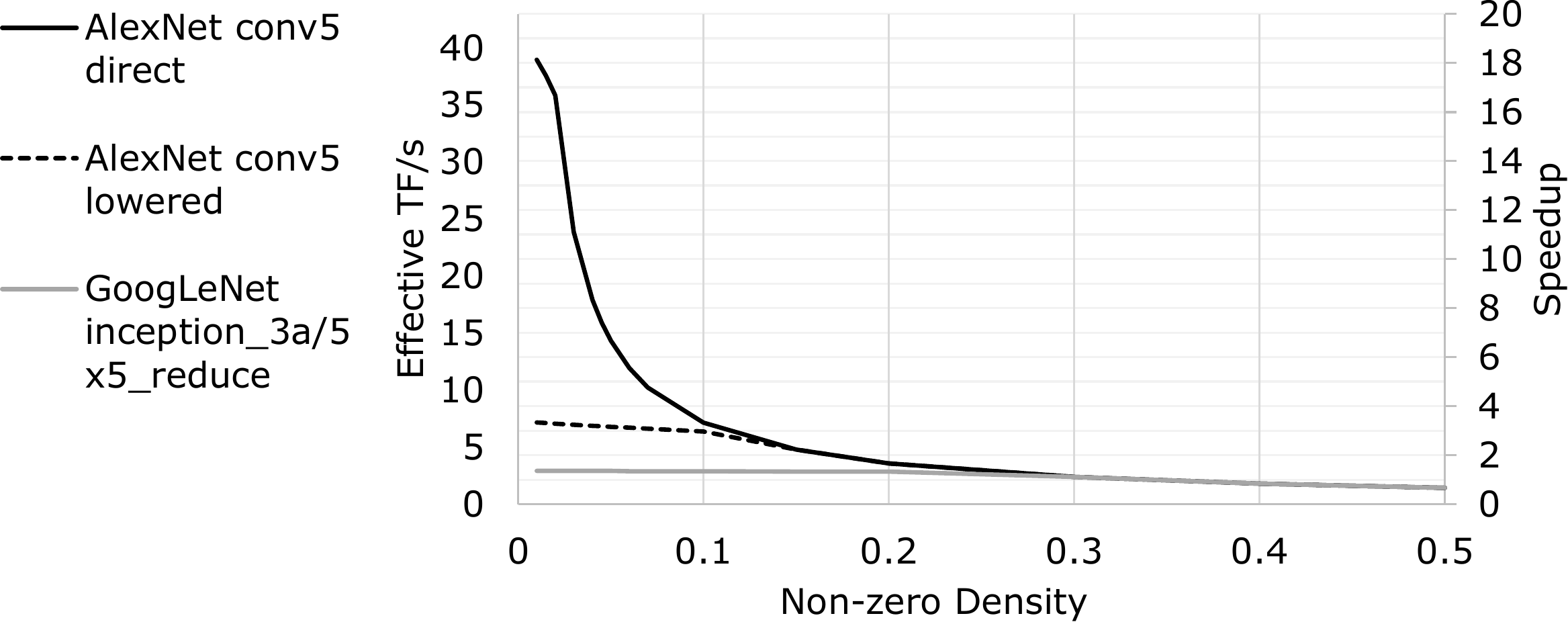}
}
\subfloat[b][Atom C2750]{
\includegraphics[width=.43\columnwidth]{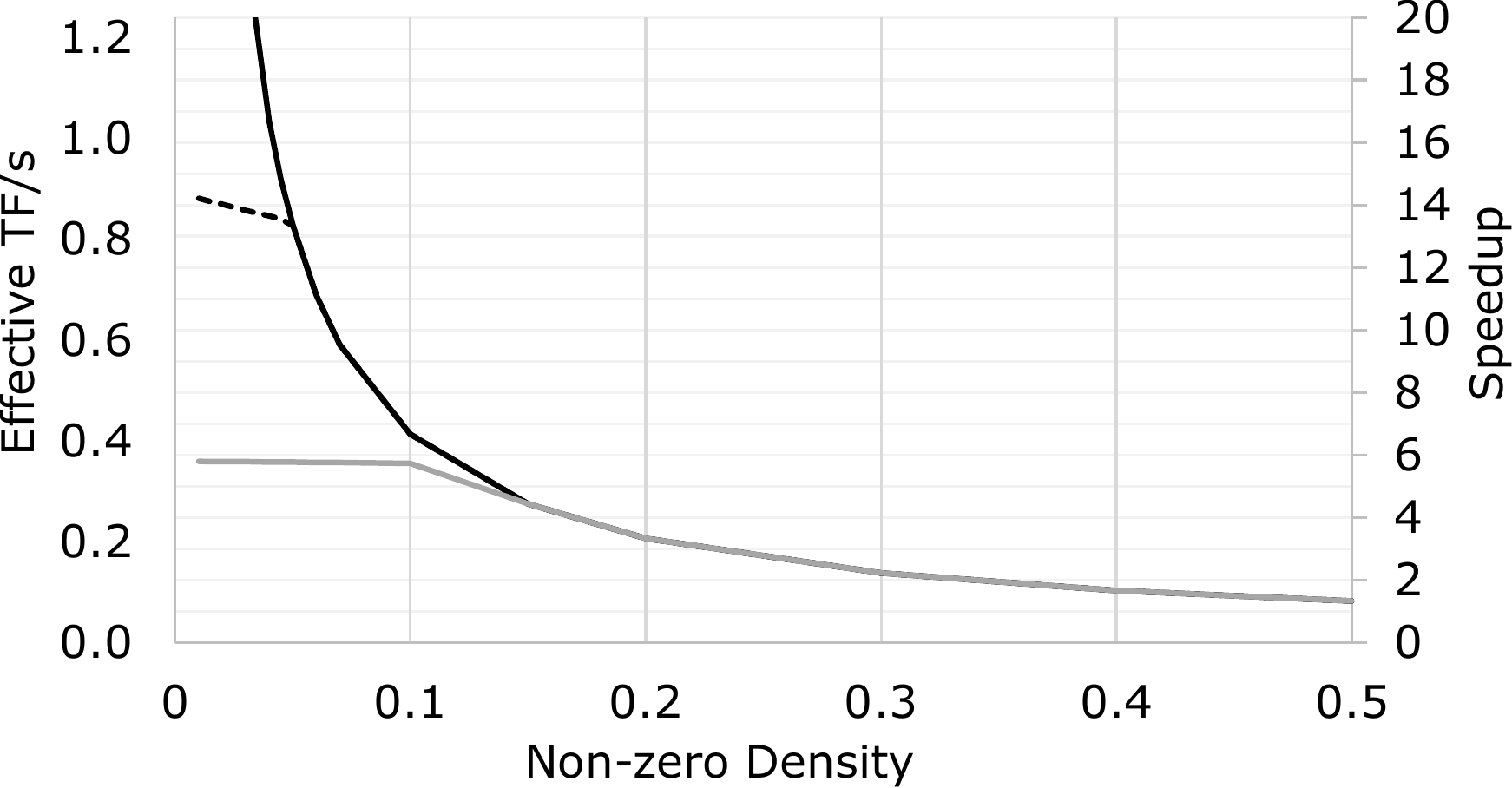}
}
\caption{\small
Projected performance of sparse convolution and its speedup over dense
convolution for a Xeon processor and an Atom processor.
%Y-axes on left and right are for relative and actual speedups, respectively.
{\tt conv5 direct}: direct spares convolution, {\tt conv5 lowered}: sparse
convolution on tensors lowered to matrices.
We use the processors' achievable FLOP/S and memory bandwidth shown in Table~\ref{tab:setup} and the compute overhead of sparse convolution
measured in Section~\ref{subsec:layer-experiment}.
}
\label{fig:arithmetic-intensity}
\end{figure}

This analytical model is visualized in
Figure~\ref{fig:arithmetic-intensity}.
Here, we define {\em effective FLOP/s} with respect to the number of
floating-point operations that would have been performed by dense convolutions including
the ones for zero weights (i.e. effective FLOP/s = $C/t_{\text sparse}$).
With a moderate sparsity (e.g., $x=0.2$), convolution is likely to be compute
bound, and hence effective FLOP/s rapidly increases as $x$ decreases.
For {\tt conv5} in AlexNet with $x \in (0.05, 0.15)$, a typical sparsity range
without accuracy loss, direct sparse convolution can achieve 2--7$\times$ and
4--14$\times$ speedup on Xeon and Atom platforms, respectively, as shown in
Figure~\ref{fig:arithmetic-intensity} and will be validated in
Section~\ref{subsec:layer-experiment}.

However, decreasing arithmetic intensity further with lowering $x$ eventually
makes the performance bandwidth bound.
Thus, \emph{there is an upper bound of useful sparsity, and a sparsity
higher than it does not provide additional speedup, while only making training
more challenging to preserve accuracy.}
This upper bound can be found by solving for $x$ such that
$t_\text{sparse\_compute} = t_\text{sparse\_bw}$ (e.g., the upper bound
sparsity for {\tt conv5} of AlexNet on the Xeon is $x\sim 0.02$).
This analysis can be applied to various computing platforms including CPUs and
GPUs because the model captures the essential platform-dependent
characteristic, the ratio of bandwidth compute capability to memory bandwidth
($F/B$).
When the compute to bandwidth ratio is lower as in a platform like Atom, the
performance will be less quickly bandwidth bound.
For example, the lower bound of useful sparsity for {\tt conv5} of AlexNet is
$x\sim 0.01$ on Atom C2750, which is smaller than that of Xeon.
The speedup to sparsity relation also varies over layers.
For example, since 1$\times$1 convolutions in GoogLeNet has low arithmetic
intensity to begin with, its performance quickly becomes bandwidth bound at
lower sparsity (or higher $x$).

The compute overhead, $\alpha$, depends on the quality of sparse
convolution implementation and on the target processor architecture.
Since $t_\text{sparse\_compute} > t_\text{dense}$ for $x > 1/\alpha$, {\it there is a
lower bound of useful sparsity such that, with a sparsity lower than that,
sparse convolution becomes slower than dense convolution}.
The previous section described our sparse convolution implementation that
achieves $\alpha$=3 (since $\alpha$ is the compute overhead, lower is better) on the Xeon
instead of $\alpha$=100 as conjectured by \cite{googlenet}\footnote{\scriptsize
The compute overhead of $\alpha$=3 primarily comes from that access to input
tensor is not aligned at cache line boundaries.
Recent Xeon processors can execute 1 unaligned SIMD load per cycle, which is
not enough to sustain 2 SIMD fused multiply-add operations per cycle.
In addition to this 2$\times$ overhead, when $W_{out}$ is not a multiple of
SIMD width (8 for Xeon E5-2697 v4), we do not fully utilize the SIMD registers.
Since Atom processors do not execute multiple SIMD floating operations per
cycle anyway, and because its SIMD width is narrower as 4, its compute overhead
is smaller as 1.2 as will be shown in Section~\ref{subsec:layer-experiment}.
}.

\subsection{Guided Sparsity Learning (GSL)}
The upper and lower bounds on useful sparsity can provide important insights for training/pruning.
The model can tell that sparse convolution is not useful for certain layers, in
which case we can skip pruning of those layers to provide more room for sparsity
in the other layers.
For example, layers like the first layer in AlexNet and GoogLeNet  may not
provide enough sparsity regardless of the amount of regularization applied
as long as the original inference accuracy is to be preserved.
A layer may be already bandwidth bound even before pruning like 1$\times$1
convolution layers in GoogLeNet as shown by {\tt inception\_4a/5x5\_reduce}
layer in Figure~\ref{fig:arithmetic-intensity}.

Guided Sparsity Learning (GSL), our new pruning algorithm, is inspired by the insights and our performance model. GSL is the \emph{first} to fuse the awareness of speedup potential into sparsity learning. GSL is a generic algorithm and accepts different regularization methods for pruning. When GSL is used with \emph{element-wise} regularization for pruning, thus denoted as Guided Element-wise Sparsity Learning (GESL),  it learns the element-wise sparsity of layers where the model predicts speedups.
%the network to materialize the full potential to accelerate inference. %When GSL is paired with \emph{group-wise} regularization, thus denoted as Guided Group-wise Sparsity Learning (GGSL), it learns the group-wise sparsity of the network. Note that GGSL needs to use a performance model for group-wise sparsity instead of element-wise sparsity to correctly project speedup and the useful sparsity ranges during training/pruning. And our performance model accommodates
%both element-wise and group-wise sparsity by considering the compute and storage overheads for each case,
%as demonstrated in \ref{subsec:model}.
GSL can also be used with regularization methods that are more complicated than basic ridge and lasso regularization.
For example, GSL can be combined with dynamic network surgery~\citep{dns}, as will be shown in Section~\ref{subsec:training}
\begin{table}[!h]
 \centering
 \footnotesize
 \newcolumntype{P}[1]{>{\raggedright\arraybackslash}p{#1}}
 \setlength\tabcolsep{1pt}
 \begin{tabular}{>{\bfseries}rP{12.5cm}}
 \hline
 & \multicolumn{1}{c}{\bf Algorithm:  Guided Sparsity Learning (GSL)}\\
 \hline
 Input: & Pruning layer set ($S$), performance model ($M$)\\%, \emph{Optional} constraints on tradeoffs among inferencing speed, accuracy, and model size\\
Initialize & Project speedup for each layer $L$ in $S$ using $M$; Exclude all $L$s without speedup potential from $S$\\
%\hspace{0.5cm} initialize pruning hyperparameters\\
Repeat & Train the whole neural network while actively pruning only $L$s in $S$ \\
%& At each interval:\\
& Project speedup of $L$s in $S$, using their current sparsity and $M$, periodically\\
& {Periodically, for each $L$ in $S$ do:} \\\
& \hspace{0.2cm} if (sparsity $\geq$ upper bound of the useful sparsity range), stop pruning $L$\\
& \hspace{0.2cm} if (stabilized sparsity $\leq$ lower bound), stop pruning $L$ \& restore its original dense weights\\
& \hspace{0.2cm} /$\ast$ Stop pruning $L$ is to give other $L$s better chance to prune further and achieve better accuracy $\ast$/\\
Until & Maximum iterations or convergence reached\\
\hline
 \end{tabular}
 %\caption*{ \label{tab:GESL_ALG}}
 \end{table}

Although GSL as described above aims primarily at
inference speed, %\footnote{In this mode, it stops pruning a layer once its sparsity reaches upper bound so that other layers will have better chance to be pruned further, although continuing pruning the layer may further reduce its size without further speedup.},
GSL can balance the implications of pruning on inference speed, accuracy, and model size. To do this, optional constraints can be given to GSL to prioritize pruning of different layers in the network. For example, by using different regularization strengths on {\tt conv} and {\tt fc}, we can tune the priorities on speed and model size. %More information about balancing accuracy, inference speed, and model size can be found in our technical report~\citep{holistic_sparse_cnn}.%For example, each {\tt conv} and {tt fc} layer has a multiplier (default value as one) to denote its relative importance during pruning, and thus larger multipliers on {\tt conv} than on {tt fc} layers instruct GESL to prioritize inferencing speed over model size, and vise versa.
% works in the following steps: 1) for a given CNN architecture, GSEL applies the analytical performance model as decribed in Section~\ref{subsec:perfmodel} and exclude the layers without\sli{No waist since it know the ranges. and precisely target the layers with high potential to achieve the goal}

%% file: experiments.tex
\section{Experiments}
\label{sec:experiments}

\definecolor{Gray}{gray}{.9}
\begin{table}[b]
\centering
\caption{\small Evaluated Platforms}
\label{tab:setup}
{\footnotesize
\begin{tabular}{r r r r}
\hline
\rowcolor{Gray}
                                        & Atom C2750 ({\tt Atom}) & Xeon E5-2697 v4 ({\tt BDW}) & Xeon Phi 7250 ({\tt KNL}) \\
Socket$\times$core$\times${\sc sp-simd} & 1$\times$8$\times$4     & 2$\times$18$\times$8        & 1$\times$68$\times$16 \\
%Memory ({\sc gb}) & 64 & 16 \\
Clock ({\sc gh}z)                       & 2.4                     & 2.3                         & 1.4                       \\
%L1/L2/L3 Cache ({\sc kb}) & 32/256/46,080$^*$ & 24/1,024$^\#$/No-L3 \\
{\tt SGEMM} {\sc gflop}/s                     & 62                      & 2,150                       & 4,540                     \\
Achievable bandwidth ({\sc gb}/s)       & 15                      & 122                         & 480                       \\
\hline
\end{tabular}
}
\begin{scriptsize}
\\
%$^*$shared among all cores in a socket, $^\#$shared among 2 cores
\end{scriptsize}
\end{table}

Our sparse CNN design is evaluated on three platforms shown in
Table~\ref{tab:setup}.
Intel C2750 ({\tt Atom}) represents resource-constrained mobile platforms or micro
servers optimized for energy efficiency.
Xeon E5-2697 v4 ({\tt BDW}) represents data-center servers.
Xeon Phi 7250 ({\tt KNL}) is designed for high-performance computing, but
its next version, Knights Mill, will specifically target machine learning.
Our sparse CNN is implemented as an extension of Caffe deep learning
framework~\citep{caffe} and is at https://github.com/IntelLabs/SkimCaffe.
We use Intel compiler version 17.0.0 and use all cores available.
The {\tt SGEMM} performance and achievable memory bandwidth listed are measured
with Intel MKL version 2017 and {\tt STREAM} benchmark~\citep{stream}, respectively.

We train with the ImageNet ILSVRC-2012 dataset~\citep{imagenet},
starting from the pre-trained Caffe reference model (a slight variation but we
call it AlexNet for simplicity) and GoogLeNet model from the Caffe model zoo.
Since we find it is easy to have high sparsity with smaller networks and
datasets like LeNet and CIFAR regardless of pruning method, we do not present
their results.
Our training process is based on the method described in \cite{sscnn} with the
following differences.
We look for element-wise sparsity with lasso instead of group lasso,
and guide the training process to target the layers and range of sparsity where
we see speedup potential.
We have explored various solver methods and learning rate schedules, but found
that they do not significantly affect the eventual accuracy and sparsity, once
hyper-parameters are tuned for the respective settings.
In general, the pruning step no longer improves after 450K and 900K mini-batch
iterations for AlexNet and GoogLeNet, respectively.
The re-training step saturates around 150K and 300K mini-batch iterations.
To see trade-offs among accuracy, speed, and model size, we try various weight
decays ranging from 1e-5 to 1e-3, and, for AlexNet, decay multipliers for {\tt
fc} layer ranging from 1e-2 to 1.
We find that the starting learning rate of 1e-3 and weight decay of 5e-5 in
general gives a high sparsity with minimal accuracy drop.
We reduce the learning rate by 10$\times$ for re-training step.

\subsection{Guided Training Results}
\label{subsec:training}

\iffalse
\begin{table}
\centering
\caption{\small Overall FLOP and non-zero reduction of
%{\tt L1'} in AlexNet uses 10 times smaller regularization on {\tt fc} layers
%to prioritize inference speed over model size, and it also does not prune {\tt
%conv1} because there is not enough sparsity to get speedups anyway.
%{\tt L1'} in GoogLeNet does not prune most of 1$\times$1 convolutions whose
%performance is memory-bandwidth bound and do not get much benefit from sparse
%convolutions, and it also does not prune the first layer (7$\times$7
%convolution) due to the same reason as {\tt conv1} in AlexNet.
}
\label{tab:overall-train}
{\footnotesize
\begin{tabular}{c | c c c c | c c c}
\hline
\rowcolor{Gray}
                    & \multicolumn{4}{c|}{AlexNet}                          & \multicolumn{3}{c}{GoogLeNet} \\
                    & L1          & L1'         & DNS          & SSL         & L1           & L1'          & SSL \\
\hline
Top-1 accuracy      & 57.2\%      & 57.4\%      &  56.9\%      & 57.5\%      & 66.7\%       & 66.7\%       & 66.9\% \\
FLOP Reduction      & 2.3$\times$ & 3.6$\times$ &  2.8$\times$ & 1.3$\times$ & 4.3$\times$  &  2.6$\times$ & 2.5$\times$ \\
Non-zeros Reduction & 9.9$\times$ & 5.7$\times$ & 17.7$\times$ & 1.0$\times$ & 6.2$\times$  &  3.3$\times$ & 2.1$\times$ \\
\hline
\end{tabular}
}
\end{table}
\fi

\begin{figure}
\subfloat[b][AlexNet]{
\includegraphics[width=.28\columnwidth]{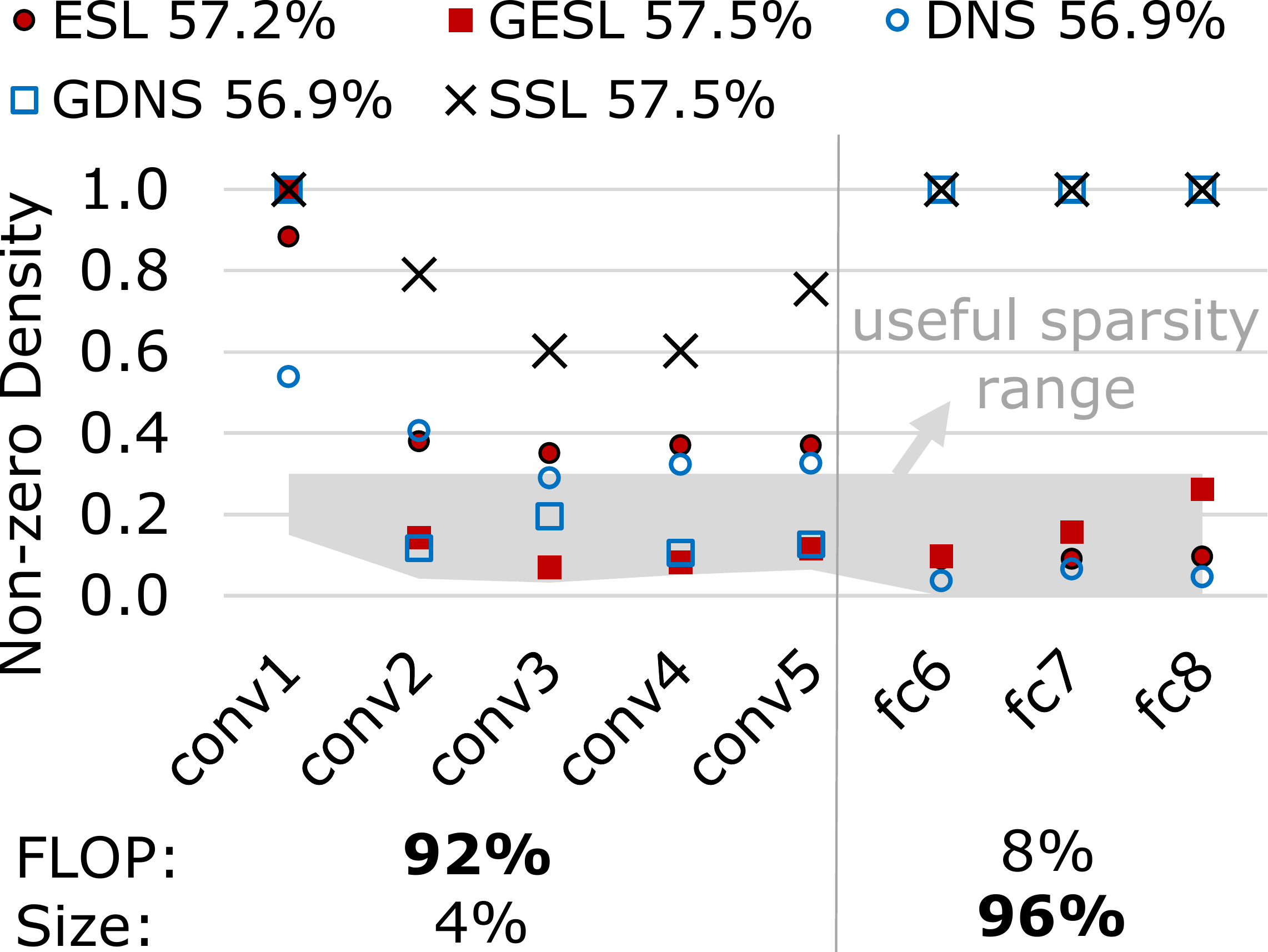}
}
\hfill
\subfloat[b][GoogLeNet]{
\includegraphics[width=.71\columnwidth]{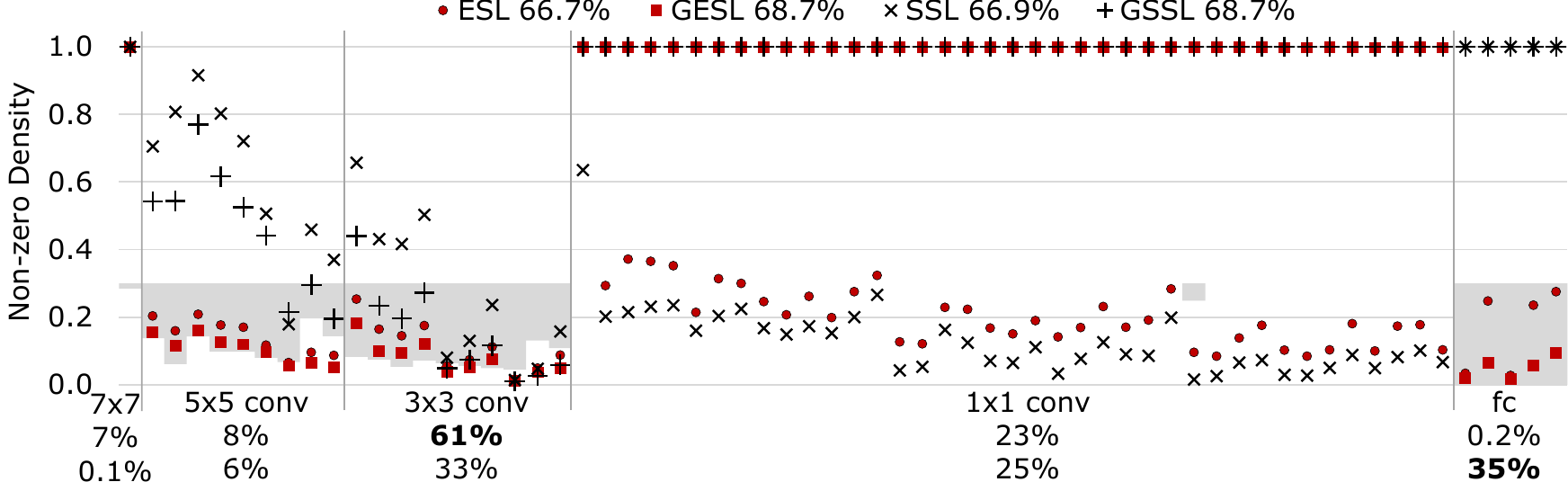}
}
\caption{\small
Layer-by-layer sparsity from element-wise sparsity learning (ESL), guided ESL,
dynamic network surgery (DNS), guided DNS, and structured sparsity learning
(SSL).
The accuracies shown in percentage are top-1 accuracy measured with the
ImageNet test set.
The original AlexNet and GoogLeNet top-1 accuracies are 57.4\% and 68.9\%.
DNS and SSL AlexNet results are from \cite{dns} and \cite{sscnn}.
GDNS AlxeNet and SSL GoogLeNet results are our own but with the same code used
in their papers.
%In both CNNs, GESL prioritizes sparsity in layers with more FLOPs and where our
%performance model projects speedups from sparse convolutions.
The shaded area marks the useful sparsity range predicted by our model for {\tt
BDW}.
No shaded area means sparse convolution is not useful for the layer regardless
of sparsity.
%GESL AlexNet results similary prioritze layers but we go further not pruning
%{\tt fc} layers at all to see how much sparsity we can get in {\tt conv} layers
%when we are solely optimizing for inference speed.
In GoogLeNet, we organize layers by their types, and, within each layer type,
layers are ordered from the earlier layers in forward propagation.
}
\label{fig:layer-by-layer-train}
\end{figure}

Figure~\ref{fig:layer-by-layer-train} shows the effectiveness of our guided
pruning and compares the level of element-wise and group-wise sparsity we can
obtain.
{\em We should look at layer-by-layer because the speedup over
dense convolution does not have a simple linear relation with sparsity as shown
by our model, and, therefore, the overall FLOP reduction does not necessarily
closely correlate with the real speedup.}
%The following section will also show that sparse convolution can speedup when
%non-zero density is smaller than 0.3.
In AlexNet, using the same element-wise regularization factor across all layers
(element-wise sparsity learning, ESL) provides non-zero densities around 0.4
for {\tt conv2-5}.
This is fine sparsity when the primary goal is reducing model size, but not
high enough for speeding-up inference.
Therefore, guided ESL (GESL) reduces the regularization factor of {\tt fc}
layers (as they have fewer FLOPS) and avoid pruning {\tt conv1} entirely (as
its sparsity is too low for any potential speedups with more regularization).
This leads to less than 0.2 non-zero density for {\tt conv2-5}, the range
where we can get speedups from sparse convolution.
Similarly, applying GSL to dynamic network surgery (DNS), a recent proposal to obtain high sparsity, as Guided DNS (GDNS), we can see that GSL effectively improve the obtained sparsity for accelerating inference by de-prioritizing {\tt
conv1} and {\tt fc} layers (we go further to not prune {\tt fc} layers at all
to see how much sparsity DNS can provide in {\tt conv} layers)\footnote{\scriptsize Although not shown in Figure~\ref{fig:layer-by-layer-train}(b), we also apply DNS and GDNS to GoogLeNet. Compared to DNS, GDNS on GoogLeNet successfully reduces non-zero density by 1.4$\times$ on average in layers with speedup potential by prioritizing these layers for pruning.}.

Structured sparsity learning (SSL) provides group-wise sparsity, for which we
can use dense methods, but its sparsity is lower because of constrained
forms of sparsity.
According to our model, SSL performs better when $x_g < (\alpha/\alpha_g)x$,
where $x$ and $x_g$ are non-zero density of ESL and SSL, and $\alpha$ and
$\alpha_g$ are the compute overhead of ESL and SSL, respectively.
Even if we use an {\em ideal} 100\% efficiency for SSL ($\alpha_g =
1$)\footnote{\scriptsize
Note that certain kinds of group-wise sparsity like ``column-wise sparsity''
defined in \cite{sscnn} need lowering, which can be considerable
overhead, making it hard to approach the ideal efficiency.}
and the {\em measured} overhead $\alpha = 3$ for ESL, $x_g$ shown in
Figure~\ref{fig:layer-by-layer-train}(a) is not small enough to outperform
GESL.
Note that our guiding principles are already applied to SSL, where {\tt conv1}
and {\tt fc} layers are not pruned.
In short, sparsity SSL can currently obtain is too low to outperform
once compared with an optimized sparse convolution design for element-wise
  sparsity such as ours.
This motivates further investigation of pruning methods for higher group-wise
sparsity.

GoogLeNet has many 1$\times$1 convolutions, for which sparse convolution does
not provide speedups due to their low arithmetic intensity, as our model
predicts.
As shown in Figure~\ref{fig:layer-by-layer-train}(a), GESL successfully discovers this and avoids pruning the 1$\times$1 convolutions
for higher sparsity in 3$\times$3 and 5$\times$5 convolutions, where our model
projects speedup, and, more importantly, almost recovers the original accuracy.
For 1$\times$1 convolutions, group-wise sparsity implemented in SSL reduces to
element-wise sparsity\footnote{\scriptsize
This is because filter coefficients for a given input and output channel pair
is also a type of group that SSL is looking for.}
%We believe SSL provides higher sparsity for 1$\times$1 convolutions because SSL
%did not prune {\tt fc} layers, providing more room to prune other layers.
%However, no pruning in {\tt fc} layers is not an inherent limitation of SSL.
%This is just because \cite{sscnn} focus on {\tt conv} layer, and we also want
%to see maximum sparsity SSL can get in {\tt conv} layers similarly to our GDNS
%AlexNet experiments.}
, and dense methods can no longer be used. We believe that SSL provides higher sparsity for 1$\times$1 convolutions than ESL because SSL
does not prune {\tt fc} layers, providing more room to prune other layers\footnote{\scriptsize However, no pruning in {\tt fc} layers is not an inherent limitation of SSL.
This is just because \cite{sscnn} focus on {\tt conv} layer, and we follow the same approach
to see maximum sparsity that SSL and GSSL can get in {\tt conv} layers.}.
For larger convolutions that contribute to the bulk of FLOPs,
ESL provides significantly higher sparsity than SSL; most of the larger
convolution layers have non-zero density less than 0.2, where sparse
convolutions can provide speedups.
%One can observe GESL's further potential to improve the accuracy with
%less regularization on a few 3$\times$3 and 5$\times$5 convolution layers whose
%sparsity exceeds the useful upper bound.
It is interesting to note that ESL, GESL, and SSL all achieve very high
sparsity of non-zero density less than 1.5\% in layers like {\tt
inception\_4e/3x3}.
This may indicate that the 3$\times$3 path is unnecessary in that inception
module.

\subsection{Layer-by-Layer Sparse Convolution Results}
\label{subsec:layer-experiment}

\begin{figure}
\subfloat[b][{\tt Atom}]{
\includegraphics[width=.32\columnwidth]{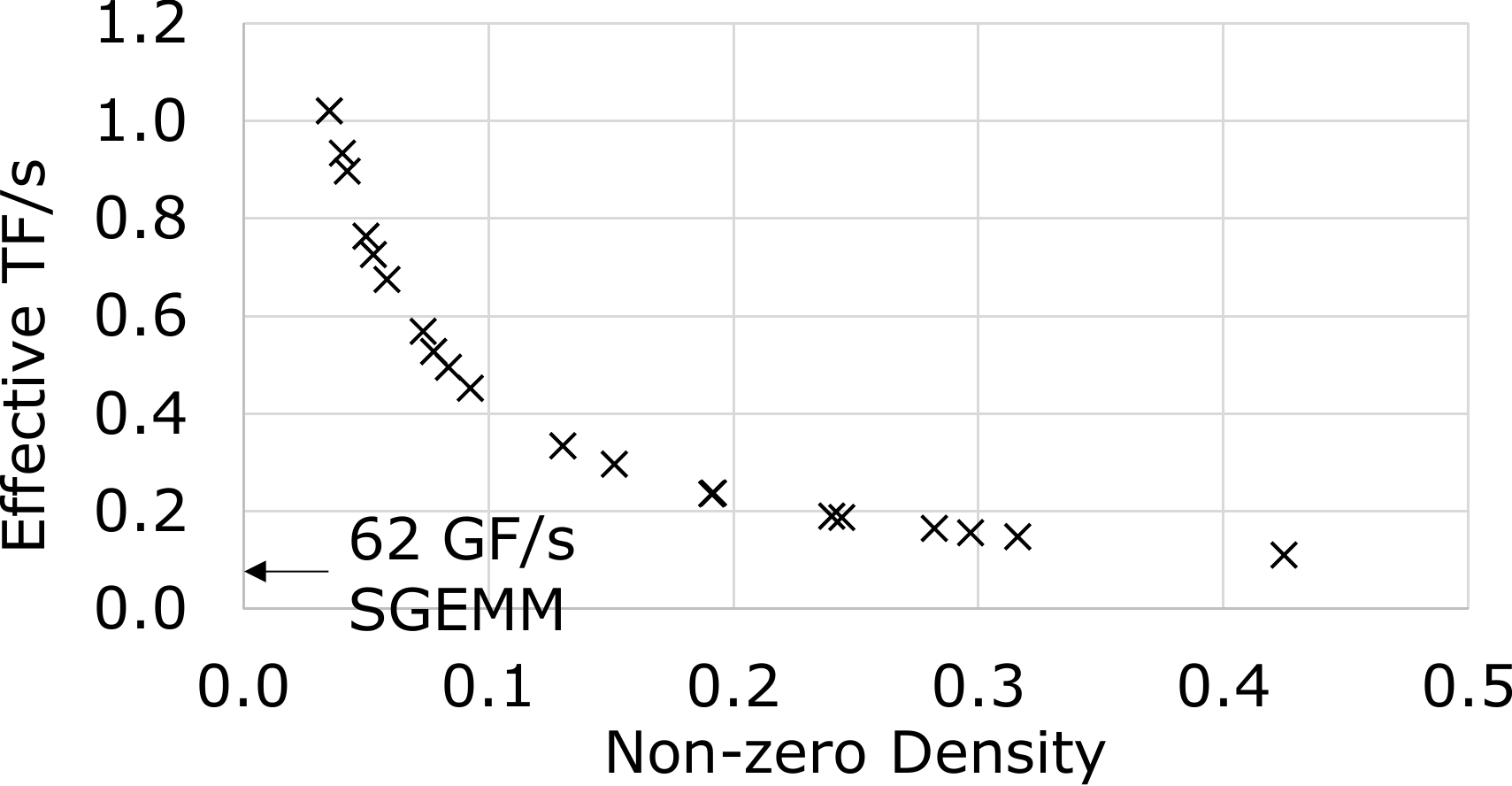}
}
\hfill
\subfloat[b][{\tt BDW}]{
\includegraphics[width=.32\columnwidth]{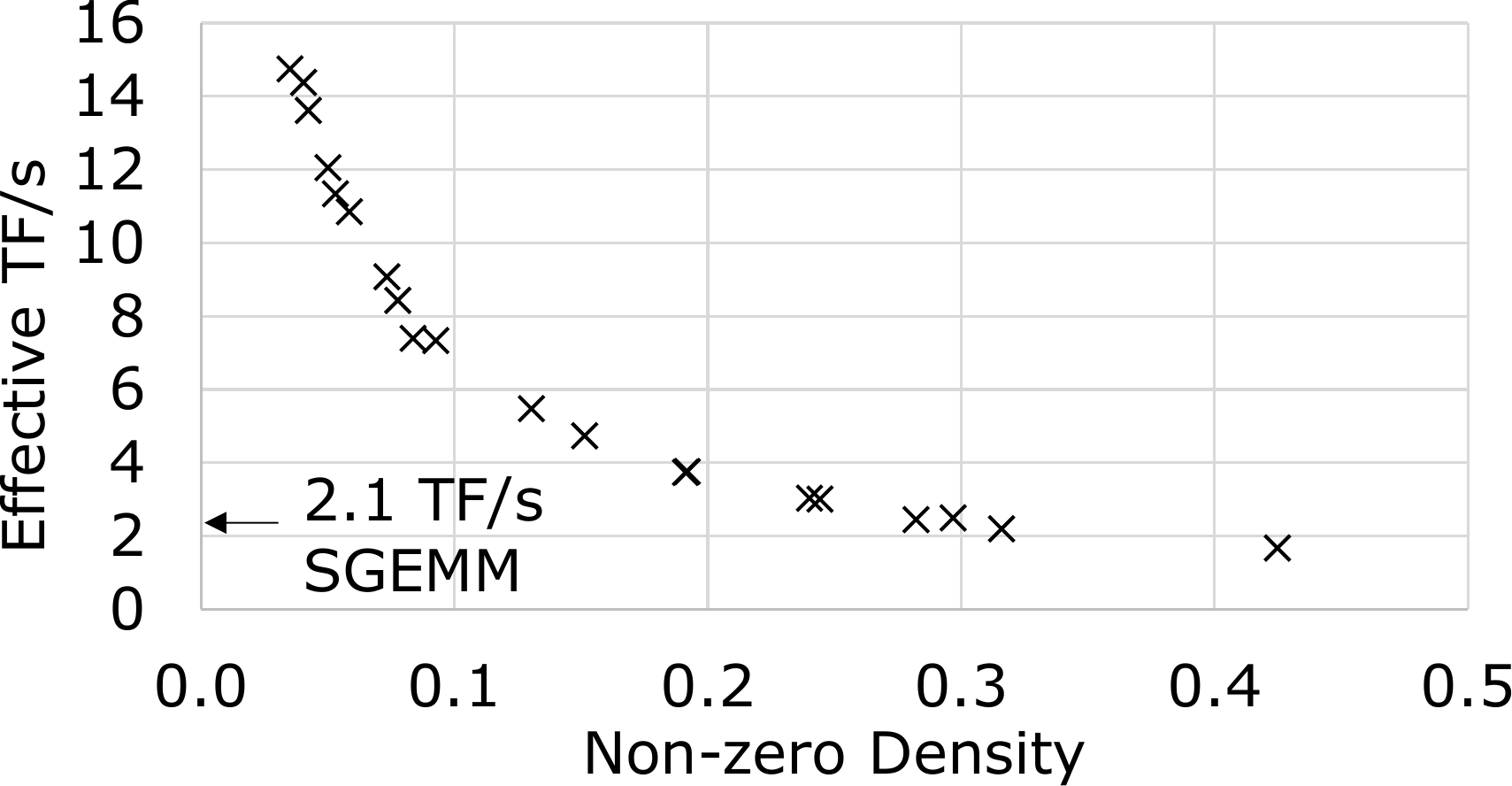}
}
\hfill
\subfloat[b][{\tt KNL}]{
\includegraphics[width=.32\columnwidth]{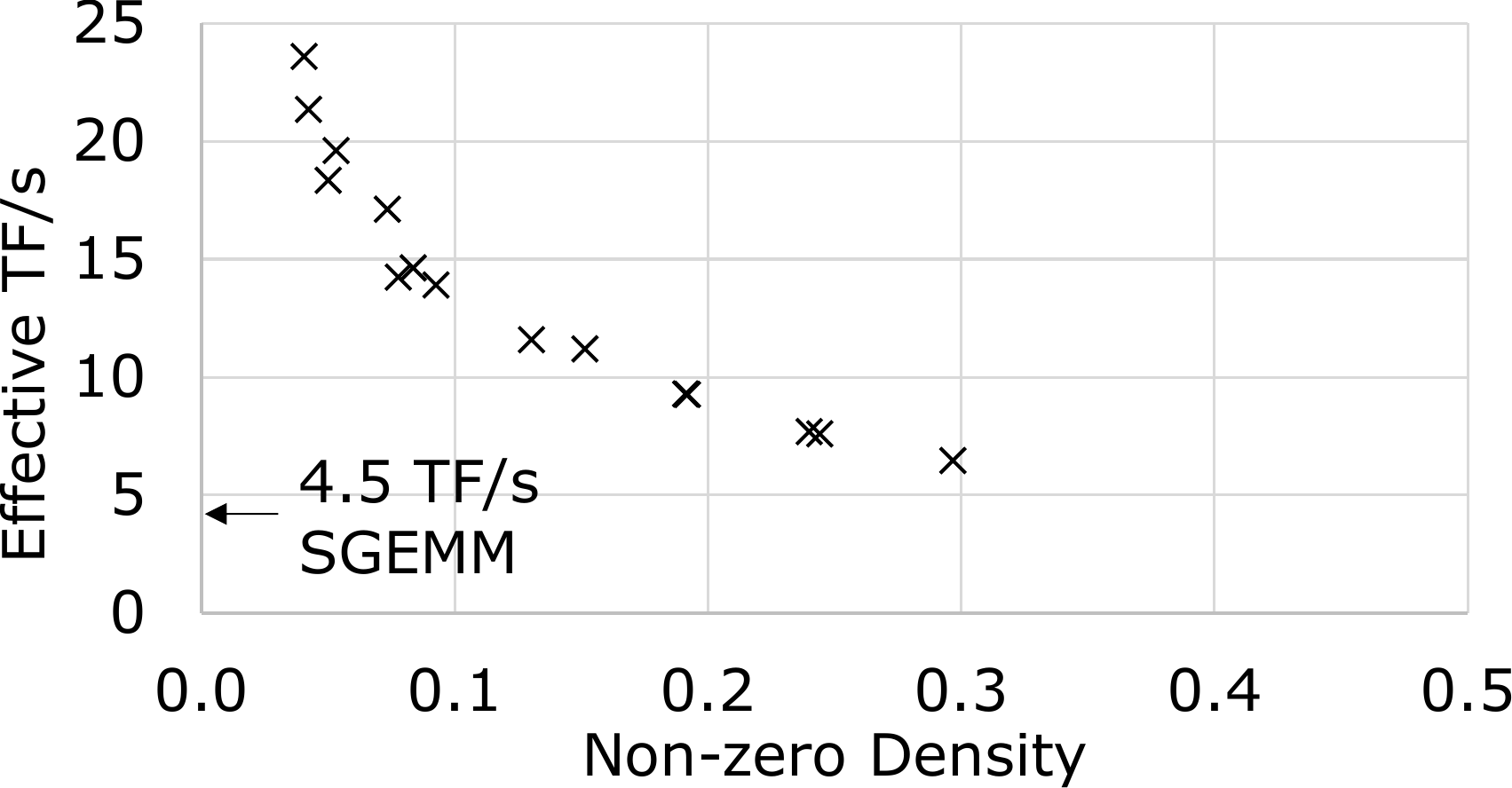}
}
\caption{\small
Performance of {\tt conv2-5} layers of AlexNet with varying sparsity
on Atom C2750 (a), Xeon E5-2697 v4 (b), and Xeon Phi 7250 (c).
{\tt SGEMM} performance of each platform serves as a proxy to the performance
of dense convolutions.
}
\label{fig:layer-wise}
\end{figure}

Figure~\ref{fig:layer-wise} shows layer-wise performance of our direct sparse
convolution design with the useful high sparsity obtained from GESL.
We evaluate with the sparse matrices from multiple pruned AlexNet models with
up to 3\% top-1 accuracy drop.
Since the performance of sparse matrix operations highly depends on specific
sparsity pattern, it is important not to evaluate with random sparse matrices.
%Similar to our projection of {\tt conv1} performance with not enough sparsity,
We use {\tt SGEMM} as a proxy of dense convolution performance to
quantify layer-wise speedups of direct sparse convolution.
{\tt SGEMM} is a good proxy because it has a long history of
extensive optimizations, and it allows us not to depend on the quality of a
specific dense convolution implementation\footnote{\scriptsize This is the same reason for this paper to focus on layer-wise performance instead of overall end-to-end speedup. As the baseline for overall end-to-end speedup may be relative to a baseline whose efficiency is suboptimal with performance bottlenecks in other parts/layers of the code. For more scientific comparison among different CNN speedup techniques, we recommend using dense matrix multiplication (GEMM) FLOP/s of the evaluated platform as the baseline, because many platforms readily have vendor-provided extensively-optimized GEMM implementations which can be a proxy of highly-optimized dense CNN implementation. This also aligns with a long-accepted standard practice in high performance computing community.}.
We use batch sizes of 32, 144, and 272 for {\tt Atom}, {\tt BDW}, and {\tt
KNL}, multiples of the number of hardware threads in respective platforms.
%Note that the baseline performance based on {\tt SGEMM} should be faster than
%any approach that uses lowering followed by {\tt SGEMM} because the overhead of
%lowering is not included.
%This is because the performance of dense convolution can widely vary depending
%on the level optimizations in a specific implementation, and not comparing with
%the best optimized dense convolution can amplify the speedup of sparse CNN.
%For example, Caffe running on CPUs involves overhead of lowering-lifting and
%does not exploit parallelism across multiple images within a mini
%batch~\cite{caffe-con-troll}.

{\tt BDW} achieves 3.4$\times$ speedup with non-zero density $x$ = 0.09, the
sparsity similar to those of {\tt conv2-5} with no accuracy drop.
The actual TF/s (as opposed to effective TF/s that also counts FLOPs for zeros)
is 0.76 when sparse convolution is sufficiently compute bound (e.g., $x>0.4$).
This performance corresponds to about a third of {\tt SGEMM}, from which we can
derive the compute overhead of sparse convolution $\alpha$ as 3.
As explained in Section~\ref{subsec:model}, this leads to the lower-bound of sparsity
to get speedups at $x=0.3$, which matches with Figure~\ref{fig:layer-wise}(b).
%This minimum sparsity to obtain speedups guides our training process to get the
%results shown in Figure~\ref{fig:layer-by-layer-train}.
{\tt Atom} with a higher bandwidth to flop ratio achieves higher 7.3$\times$
speedup at $x=0.09$.
The actual GF/s is 51 when $x>0.4$, which is 1.2$\times$ lower than {\tt SGEMM}
performance (i.e. $\alpha = 1.2$).
Note that the performance projection for {\tt conv5} in
Figure~\ref{fig:arithmetic-intensity} using $\alpha$s derived here resembles
the measured performance in Figure~\ref{fig:layer-wise} ({\tt conv2-5} share
similar performance characteristics).
{\tt KNL} achieves impressive 13.9 effective TF/s at $x=0.09$ (3.1$\times$ over
{\tt SGEMM}).

%Plugging in the layer-by-layer performance to the training result of GESL on
%AlexNet shown in Figure~\ref{fig:layer-by-layer-train}(a), we estimate 2,371
%and 120 images per second end-to-end inference throughput on {\tt BDW} and {\tt
%Atom}, respectively.
%The {\tt BDW} performance is is significantly faster than the 1,420 images per
%second reported with a highly optimized CNN implementation with dense
%convolution running on a platform with a comparable compute
%platform~\citep{mkl-dnn}.
%Note that this result is with no accuracy drop, and, if we can tolerate about
%3\% accuracy loss, we estimate up to 3,612 images per second on {\tt BDW}.
%We do not project the end-to-end performance of GoogLeNet because other layers
%such as pooling and normalization account for a considerable amount of time in
%GoogLeNet unlike AlexNet.

\iffalse
As discussed in Section~\ref{sec:sparseCNN}, we also considered SpGEMM on fc
layers, but with performance worse than SpMDM.
{\bf TODO: discussion on SpGEMM performance. Or take it out if we're out of space.}
\fi

%A figure of speed up vs. Accuracy loss for each layer

%Mention size reduction, but do not need figure.

%Different scoring systems: from Server to Mobile, to show different performance impact on sparse convolution and SpMDM/Spgemm on FC layers.

%% file: related.tex
\section{Related Work}
\newcommand{\thickhline}{\noalign{\hrule height 0.8pt}}
\label{sec:related}
\newcolumntype{P}[1]{>{\centering\arraybackslash}p{#1}}
\begin{table}[!hbt]
 \centering
 \footnotesize
 \begin{tabular}{c| P{2cm} |P{3cm} | P{5.5cm}}
 %\multicolumn{3}{c}{Pruning Methods for Eliminating Redundant Parameters} \\
 & A: \cite{scnn_lebedev}\textsuperscript{$\ast$}, \cite{sscnn}\textsuperscript{$\ast$}
 & B: \cite{prune}, \cite{deep-compression}, \cite{scnn}\textsuperscript{$\ast$}, \cite{dns}, GESL\textsuperscript{$\ast$}
 & C: \cite{fact_lecun}, \cite{fact_Jaderberg}, \cite{fact_lebedev}, \cite{fact_zhang}, \cite{fact_kim}, \cite{fact_ioannou}, \cite{fact_tai}, \cite{prune_13}\\
  \thickhline
 \multirow{2}{*}{Pruning} & \multicolumn{2}{c|}{Regularization} &\multirow{2}{*}{Factorization}\\
 \cline{2-3}
 &\multicolumn{1}{c|}{Group-wise} &\multicolumn{1}{c|}{Element-wise} & \\
  \thickhline
 Computing & Dense & Sparse & Dense \\
 %\multicolumn{3}{c}{Methods for Computing Convolution: Linear Algebra Operation Types}\\
 %\parbox[t]{2mm}{\multirow{3}{*}{\rotatebox[origin=c]{90}{\parbox{3cm}{Inferencing Method}}}} & \multicolumn{1}{|c|}{Sparse} &   & \citep{prune},\citep{deep-compression}, \citep{prune_13} &\\
% \cline{2-5}
% & \multicolumn{1}{|c|}{Dense} & \citep{scnn_lebedev},\citep{sscnn}& &\citep{fact_lecun},~\citep{fact_Jaderberg},~\citep{fact_lebedev},~\citep{fact_zhang},~\citep{fact_kim,fact_ioannou},~\citep{fact_tai}\\%[50pt]
 \end{tabular}
 \caption{\small Design space of techniques in reducing model size and accelerating inference, categorized as 3 groups. The footer rows of the table specify the two pillars of design space: pruning methods (how the sparsity is obtained during training) and computing methods (how the the obtained sparsity during inference). For techniques using regularization for pruning, $^*$ denotes those focusing more on {\tt conv} layers than on {\tt fc} layers.
 %\\$\dagger$Factorization based techniques do not require training to pruning redundant parameters, although they can use fine-tuning to recover accuracy loss because of factorization.
 \label{tab:sparseCNN-taxo}
 }
 \end{table}

Recent researches have achieved great success on reducing model size and accelerating inference of CNNs while maintaining accuracy, exploring a large design space as shown in Table~\ref{tab:sparseCNN-taxo}.
%, these researches have explored a large design space. %with two pillars: the pruning methods (i.e., how the sparsity is obtained) and the computing methods (how the computation, especially convolutions, leverages the obtained sparsity). %; and 3) the target layers ---whether the technique focus more on fully connected layers (thus model reduction) or convolution layers (thus inference acceleration).
Regularization-based and factorization-based approaches are the two main camps.
Regularization-based approaches use a separate training step to discover and prune redundant parameters in a pre-trained model using various regularizations, including ridge~\citep{prune, deep-compression}, lasso, and group lasso~\citep{scnn, sscnn}, combined with thresholding. Factorization-based approaches use low-rank decomposition and can quickly produce compressed models without additional pruning steps. Both approaches can use a fine-tuning step to recover accuracy loss caused by model pruning.

%An important benefit for pruning parameters is to reduce model size. For this purpose, one thrust of
Researches focusing on fully connected layers~\citep{prune,deep-compression,prune_13} have achieved 10--50$\times$ model size reduction for networks such as AlexNet~\citep{alexnet}.  %that account for more than 90\% of the parameters in some networks, such as AlexNet~\citep{alexnet}, and achieved 10$\times$ to 50$\times$ model size reduction for these networks.
However, they achieved marginal inferencing speedup because fully connected layers usually account for less than 10\% of total computation in modern CNNs.
Researches in groups A and C shown in Table~\ref{tab:sparseCNN-taxo} aim at speeding up inference by focusing more on convolution layers, with most of them relying on dense methods for computing convolution.  %Among these researches, almost all of the factorization based approaches and some of the regularization base approaches rely on dense linear algebra operation for computing convolution.
While factorization-based approaches (group C) obtain smaller models in dense format naturally, regularization-based approaches (group A) need group regularization to impose group-wise sparsity. %to use dense methods to compute convolution, %so that the obtained sparse model can be cast back into a dense but smaller model for computation with dense linear algebra, by techniques such as lowering/lifting.
%even though the obtainable group-wise sparsity is usually lower than the obtainable element-wise sparsity.
Although \cite{scnn} explore sparse methods in computing convolution layers,
their approach involves lowering overhead and uses hard-coding non-zeros in
sparse matrix with full unrolling that leads to a large instruction footprint.

While our direct sparse convolution is demonstrated to achieve high speed up on
convolution when having enough sparsity, factorization-based approaches can
complement.
This is because the inherent sparsity in the first few convolution layers can
be not high enough, while factorization-based approaches can achieve speedups
there.
%Thus, further speedup can be achieved by using factorization-based approaches
%in the first a few convolution layers and direct sparse convolution in later
%layers.
\cite{scnn} also show that factorization and regularization-based approaches
can be combined.

Winograd~\citep{winograd} and FFT based algorithms~\citep{fft-cnn} also aim to
speedup convolution.
While being orthogonal, these techniques can have synergies with our direct
sparse convolution.
For example, FFT based convolutions are more effective for large filters that
usually reside in the first few layers where sparsity is low.
While this paper focuses on convolution layer performance, our technical
report~\citep{holistic_sparse_cnn} also considers optimizations for fully
connected layers, and sparsity in activations, which is also discussed in
\cite{EIE}.

%% file: conclusion.tex
\section{Conclusions}
\label{sec:conclusion}

Powerful CNNs are often quite compute demanding. Pruning as a post-processing
step has been effective in drastically reducing the model size while boosting
inference speed moderately.
We aim to more fully realize the potential performance benefits due to the reduced FLOP counts
resulting from pruned convolution kernels. By combining our high-performance direct sparse
convolution method with a performance model, we developed a guided approach that
prunes CNNs in a co-design fashion for different computer architectures and on
different layers of a CNN in question. In particular, we demonstrated
3.1--7.3$\times$ convolution speedups in AlexNet on a variety of platforms, all
in comparison to extensively-optimized dense linear algebra operations.

Looking ahead, as this paper shows that pruning can boost inference speed
significantly in additional to reducing model size, further techniques in
pruning should be explored. 
While our direct sparse convolution algorithm is successful, our performance
model also reveals that sparse convolution cannot speedup all convolution
layers, as seen from 1$\times$1 convolutions in GoogLeNet.
We plan to expand our performance model to cover other FLOP-reduction methods
such as FFT, Winograd, and tensor factorization, so that we can make informed
decisions to choose the best performing method for each layer and the training
process can be guided accordingly.